\documentclass[10pt,letterpaper]{article}
\usepackage[top=0.85in,left=2.75in,footskip=0.75in,marginparwidth=2in]{geometry}

\usepackage[utf8]{inputenc}

\usepackage{cite}

\usepackage{nameref,hyperref}

\usepackage[right]{lineno}

\usepackage{microtype}
\DisableLigatures[f]{encoding = *, family = * }

\raggedright
\usepackage{changepage}

\usepackage[aboveskip=1pt,labelfont=bf,labelsep=period,singlelinecheck=off]{caption}

\makeatletter
\renewcommand{\@biblabel}[1]{\quad#1.}
\makeatother

\usepackage{lastpage,fancyhdr,graphicx}
\usepackage{epstopdf}
\usepackage{ragged2e}
\justifying
\fancyheadoffset[L]{2.25in}
\fancyfootoffset[L]{2.25in}

\usepackage{color}

\definecolor{Gray}{gray}{.25}

\usepackage{graphicx}

\usepackage{sidecap}

\usepackage{wrapfig}
\usepackage[pscoord]{eso-pic}
\usepackage[fulladjust]{marginnote}
\reversemarginpar

\begin{document}
\vspace*{0.35in}

\begin{flushleft}
{\Large
\textbf\newline{Fine-tuning of Pre-trained Transformers for Hate, Offensive, and Profane Content Detection in English and Marathi}
}
\newline
\\
Anna Glazkova\textsuperscript{1,*},
Michael Kadantsev\textsuperscript{2},
Maksim Glazkov\textsuperscript{3},
\\
\bigskip
\bf{1} University of Tyumen, Tyumen, Russia
\\
\bf{2} Thales Canada, Transportation Solutions, Toronto, Canada
\\
\bf{3} Neuro.net, Nizhny Novgorod, Russia
\\
\bigskip
* a.v.glazkova@utmn.ru

\end{flushleft}

\section*{Abstract}
  This paper describes neural models developed for the Hate Speech and Offensive Content Identification in English and Indo-Aryan Languages Shared Task 2021. Our team called \textit{neuro-utmn-thales} participated in two tasks on binary and fine-grained classification of English tweets that contain hate, offensive, and profane content (English Subtasks A \& B) and one task on identification of problematic content in Marathi (Marathi Subtask A). For English subtasks, we investigate the impact of additional corpora for hate speech detection to fine-tune transformer models. We also apply a one-vs-rest approach based on Twitter-RoBERTa to discrimination between hate, profane and offensive posts. Our models ranked third in English Subtask A with the F1-score of 81.99\% and ranked second in English Subtask B with the F1-score of 65.77\%. For the Marathi tasks, we propose a system based on the Language-Agnostic BERT Sentence Embedding (LaBSE). This model achieved the second result in Marathi Subtask A obtaining an F1 of 88.08\%.


\section*{Introduction}

Social media has a greater impact on our society. Social networks give us almost limitless freedom of speech and contribute to the rapid dissemination of information. However, these positive properties often lead to unhealthy usage of social media. Thus, hate speech spreading affects users' psychological state, promotes violence, and reinforces hateful sentiments \cite{beausoleil2019free,bilewicz2020hate}. This problem attracts many scholars to apply modern technologies in order to make social media safer. The Hate Speech and Offensive Content Identification in English and Indo-Aryan Languages Shared Task (HASOC) 2021 \cite{hasoc2021mergeoverview} aims to compare and analyze existing approaches to identifying hate speech not only for English, but also for other languages. It focused on detecting hate, offensive, and profane content in tweets, and offering six subtasks. We participated in three of them:

\begin{itemize}
    \item \textbf{English Subtask A:} identifying hate, offensive, and profane content from the post in English \cite{hasoc2021overview}.
    \item \textbf{English Subtask B:} discrimination between hate, profane, and offensive posts in English.
    \item \textbf{Marathi Subtask A:} identifying hate, offensive, and profane content from the post in Marathi \cite{gaikwad2021cross}.
\end{itemize}

The source code for our models is freely available\footnote{\url{https://github.com/ixomaxip/hasoc}}.

The paper is organized as follows. Section 2 contains a brief review of related works. Next, we describe our experiments on the binary and fine-grained classification of English tweets in Section 3. In Section 4, we present our model for hate, offensive, and profane language identification in Marathi. We conclude this paper in Section 5. Finally, Section 6 contains acknowledgments.

\section{Related Works}

We briefly discuss works done related to harmful content detection in the past few years. Shared tasks related to hate speech and offensive language detection from tweets was organized as a part of some workshops and conferences, such as FIRE \cite{mandl2019overview,mandl2020overview}, SemEval,  \cite{davidson2017automated,basile2019semeval}, GermEval \cite{wiegand2018overview,struss2019overview}, IberLEF \cite{taule2021overview}, and OSACT \cite{mubarak2020overview}. The participants proposed a broad range of approaches from traditional machine learning techniques (for example, Support Vector Machines \cite{schmid2019fosil,hassan2020alt}, Random Forest \cite{ray2020ju}) to various neural architectures (Convolutional Neural Networks, CNN \cite{ribeiro2019inf}; Long Short Term Memory, LSTM \cite{mishraa2020iiit_dwd,montejo2019sinai}; Embeddings from Language Models, ELMo \cite{bojkovsky2019stufiit}; and Bidirectional Encoder Representations from Transformers, BERT \cite{risch2019hpidedis,liu2019nuli}). In most cases, BERT-based systems outperformed other approaches.

Most research on hate speech detection continues to be based on English corpora. Despite this, the harmful content is distributed in different languages. Therefore, there have been previous attempts at creating corpora and developing models for hate speech detection in common non-English languages, such as Arabic \cite{mubarak2020overview,albadi2018they}, German \cite{mandl2019overview,mandl2020overview,wiegand2018overview,struss2019overview}, Italian \cite{bosco2018overview,sanguinetti2018italian}, Spanish \cite{basile2019semeval,taule2021overview}, Hindi \cite{mandl2019overview,mandl2020overview}, Tamil and Malayalam \cite{mandl2020overview}. Several studies have focused on collecting hate speech corpora for Chinese \cite{tang2020categorizing}, Portuguese \cite{de2017offensive}, Polish \cite{ptaszynski2019results}, Turkish \cite{ccoltekin2020corpus} and Russian \cite{komalova2021corpus} languages.

\section{English Subtasks A \& B: Identification and Fine-grained Classification of Hate, Offensive, and Profane Tweets}

The objective of English Subtasks A \& B is to identify whether a tweet in English contains harmful content (Subtask A) and perform a fine-grained classification of posts into three categories, including: hate, offensive, or profane (Subtask B). 

\subsection{Data}

The dataset provided to the participants of the shared task contains 4355 manually annotated social media posts divided into training (3074) and test (1281) sets. Table 1 presents the data description.

\begin{table}[]
\caption{Data description.}
\begin{tabular}{|l|l|l|}
\hline
\multicolumn{1}{|c|}{Label} & \multicolumn{1}{c|}{Description} & \begin{tabular}[c]{@{}l@{}}Number of examples \\ (training set)\end{tabular} \\ \hline
\multicolumn{3}{|c|}{Subtask A} \\ \hline
NOT & \begin{tabular}[c]{@{}l@{}}Non Hate-Offensive: the post does not contain\\ hate speech, profane, offensive content\end{tabular} & 1102 \\ \hline
HOF & \begin{tabular}[c]{@{}l@{}}Hate and Offensive: the post contains hate, \\ offensive, or profane content.\end{tabular} & 1972 \\ \hline
\multicolumn{3}{|c|}{Subtask B} \\ \hline
NONE & \begin{tabular}[c]{@{}l@{}}The post does not contain hate speech, profane, \\ offensive content\end{tabular} & 1102 \\ \hline
HATE & Hate speech: the post contains hate speech content. & 542 \\ \hline
OFFN & Offensive: the post contains offensive content. & 482 \\ \hline
PRFN & Profane: the post contains profane words. & 948 \\ \hline
\end{tabular}
\end{table}

Further, we tested several data sampling techniques using different hate speech corpora as additional training data. Firstly, we evaluated the joint use of multilingual data provided by the organizers of HASOC 2021, including the English, the Hindi, and the Marathi training sets. Secondly, as the training sets were highly imbalanced, we applied the positive class random oversampling technique so that each training batch contained approximately the same number of samples. Besides, we experimented with the seq2seq-based data augmentation technique \cite{kumar2020data}. For this purpose, we fine-tuned the BART-base model for the denoising reconstruction task where 40\% of tokens are masked and the goal of the decoder is to reconstruct the original sequence. Since the BART model \cite{lewis2020bart} already contains the <mask> token, we use it to replace mask tokens. We generated one synthetic example for every tweet in the training set. Thus, the augmented data size is the same size as the size of the original training set. Finally, we evaluated the impact of additional training data, including: (a) the English dataset, used at HASOC 2020 \cite{mandl2020overview}; (b) HatebaseTwitter, based on the hate speech lexicon from Hatebase\footnote{\url{https://hatebase.org/}} \cite{davidson2017automated}; (c) HatEval, a dataset presented at Semeval-2019 Task 5 \cite{basile2019semeval} ; (d) Offensive Language Identification Dataset (OLID), used in the SemEval-2019 Task 6 (OffensEval) \cite{zampieri2019semeval}. All corpora except the HatebaseTwitter dataset contain non-intersective classes. Besides, all listed datasets are collected from Twitter. A representative sampling of additional data is shown in Table 2.

\begin{table}[]
\caption{Hate-related dataset characteristics.}
\begin{tabular}{|l|l|l|}
\hline
\multicolumn{1}{|c|}{Dataset} & \multicolumn{1}{c|}{Size} & \multicolumn{1}{c|}{Labels} \\ \hline
HASOC 2020 & 4522 & \begin{tabular}[c]{@{}l@{}}HOF - 50.4\%\\ NOT - 49.6\%\end{tabular} \\ \hline
HatebaseTwitter & 24783 & \begin{tabular}[c]{@{}l@{}}hate speech - 20.15\%\\ offensive language - 85.98\%\\ neither - 23.77\%\end{tabular} \\ \hline
HatEval & 13000 & \begin{tabular}[c]{@{}l@{}}1 (hate speech) - 42.08\%\\ 0 (not hate speech) - 57.92\%\end{tabular} \\ \hline
OLID & 14100 & \begin{tabular}[c]{@{}l@{}}OFF - 32.91\%\\ NOT - 67.09\%\end{tabular} \\ \hline
\end{tabular}
\end{table}

We preprocessed the datasets for Subtasks A \& B in a similar manner. Inspired by \cite{barbieri2020tweeteval}, we used the following text preprocessing technique\footnote{\url{https://pypi.org/project/tweet-preprocessor}}: (a) removed all URLs; (b) replaced all user mentions with the \$MENTION\$ placeholder. 

\subsection{Models}

We conduct our experiments with neural models based on BERT \cite{devlin2018bert} as they have achieved state-of-the-art results in harmful content detection. For example, BERT-based models proved efficient at previous HASOC shared tasks \cite{mandl2020overview,mandl2019overview} and SemEval \cite{zampieri2019semeval,pavlopoulos-etal-2021-semeval}. 

We used the following models:

\begin{itemize}
    \item BERT$_{base}$ \cite{devlin2018bert}, a pre-trained model on BookCorpus \cite{zhu2015aligning} and English Wikipedia using a masked language modeling objective.
    \item BERTweet$_{base}$ \cite{nguyen2020bertweet}, a pre-trained language model for English tweets. The corpus used to pre-train BERTweet consists of 850M English Tweets including 845M Tweets streamed from 01/2012 to 08/2019 and 5M Tweets related to the COVID-19 pandemic.
    \item Twitter-RoBERTa$_{base}$ for Hate Speech Detection \cite{barbieri2020tweeteval}, a RoBERTa$_{base}$ \cite{liu2019roberta} model trained on ~58M tweets and fine-tuned for hate speech detection with the TweetEval benchmark.
    \item LaBSE \cite{feng2020language}, a language-agnostic BERT sentence embedding model supporting 109 languages. 
\end{itemize}

\subsection{Experiments}

For both Subtask A and Subtask B, we adopted pre-trained models from HuggingFace \cite{wolf2020transformers} and fine-tuned them using PyTorch \cite{paszke2019pytorch}. We fine-tuned each pre-trained language model for 3 epochs with the learning rate of 2e-5 using the AdamW optimizer \cite{loshchilov2018decoupled}. We set batch size to 32 and maximum sequence size to 64. To validate our models during the development phase, we divided labelled data using the train and validation split in the ratio 80:20.

Table 3 shows the performance of our models on the validation subset for Subtask A in terms of macro-averaging F1-score (F1), precision (P), and recall (R). As can be seen from the table, BERT, BERTweet, and LaBSE show very close results during validation. Despite this, LaBSE jointly fine-tuned on three mixed multilingual datasets shows the highest precision score. The use of Twitter-RoBERTa increases the F1-score by 1.5-2.5\% compared to other classification models. Based on this, we chose Twitter-RoBERTa for further experiments. We found out that neither the random oversampling technique nor the use of the augmented and additional data shows a performance improvement, except the joint use of the original dataset and the HatebaseTwitter dataset that gives an F1-score growth of 0.09\% and a precision growth of 0.28\% compared to basic Twitter-RoBERTa. 

\begin{table}[]
\caption{Model validation results for English Subtask A, \%.}
\begin{tabular}{|l|l|l|l|}
\hline
\multicolumn{1}{|c|}{Model} & \multicolumn{1}{c|}{F1} & \multicolumn{1}{c|}{P} & \multicolumn{1}{c|}{R} \\ \hline
BERT & 79.24 & 79.74 & 78.82 \\ \hline
BERTweet & 78.65 & 79.36 & 78.08 \\ \hline
Twitter-RoBERTa & 81.1 & 80.01 & 82.65 \\ \hline
LaBSE (English dataset) & 78.83 & 79.5 & 78.29 \\ \hline
LaBSE (English + Hindi) & 79.32 & 79.95 & 78.8 \\ \hline
LaBSE (English, Hindi, and Marathi) & 79.27 & \textbf{81.74} & 77.79 \\ \hline
\multicolumn{4}{|c|}{Adding extra data to Twitter-RoBERTa} \\ \hline
+ random oversampling & 79.97 & 79.9 & 80.04 \\ \hline
+ BART data augmentation & 79.24 & 78.44 & 80.31 \\ \hline
+ HASOC 2020 & 78.79 & 77.66 & 80.47 \\ \hline
+ HatabaseTwitter & \textbf{81.19} & 79.99 & \textbf{82.93} \\ \hline
+ HatEval & 74.31 & 75.53 & 73.64 \\ \hline
+ OLID & 79.29 & 78.17 & 80.93 \\ \hline
\end{tabular}
\end{table}

For our official submission for Subtask A, we designed a soft-voting ensemble of five Twitter-RoBERTa jointly fine-tuned on the original training set and the HatebaseTwitter dataset (see Table 4). For Subtask B, we used the following one-vs-rest approach to discrimination between hate, profane, and offensive posts.

\begin{itemize}
    \item First, we applied our Subtask A binary models to identify non hate-offensive examples.
    \item Second, we fine-tuned three Twitter-RoBERTa binary models to delimit examples of hate-vs-profane, hate-vs-offensive, and offensive-vs-profane classes. The training dataset was extended with the HatebaseTwitter dataset.
    \item Finally, we compared the results of binary models. If the result was defined uniquely, we used it as a predicted label. Otherwise, we chose the label in proportion to the number of examples in the training set. 
    
    This can be illustrated briefly by the following examples. 
    \begin{itemize}
        \item Let the models show the following results: 
        \begin{itemize}
            \item hate-vs-profane$\rightarrow$ hate;
            \item hate-vs-offensive$\rightarrow$ hate;
            \item offensive-vs-profane$\rightarrow$ offensive. 
        \end{itemize}
        Thus, classes have the following votes: hate -- 2, offensive - 1, profane -- 0. Then we predict the HATE label. 
        \item If the results are:
        \begin{itemize}
            \item hate-vs-profane$\rightarrow$ profane;
            \item hate-vs-offensive$\rightarrow$ hate;
            \item offensive-vs-profane$\rightarrow$ offensive, 
        \end{itemize}
        we have the class votes, such as hate -- 1, offensive - 1, profane -- 1. Then we choose the PRFN label as the most common label in the training set.
    \end{itemize}
\end{itemize}

\begin{table}[]
\caption{Performance of our final models for English Subtasks A \& B, official results, \%.}
\begin{tabular}{|l|l|l|l|l|l|l|l|}
\hline
Subtask & \begin{tabular}[c]{@{}l@{}}F1 (our \\ model)\end{tabular} & \begin{tabular}[c]{@{}l@{}}F1 (winning\\ solution)\end{tabular} & \begin{tabular}[c]{@{}l@{}}P (our\\ model)\end{tabular} & \begin{tabular}[c]{@{}l@{}}P (winning\\ solution)\end{tabular} & Avg F1 & \begin{tabular}[c]{@{}l@{}}Number of\\ submitted\\ teams\end{tabular} & Rank \\ \hline
A & 81.99 & 83.05 & 84.68 & 84.14 & 75.7 & 56 & 3 \\ \hline
B & 65.77 & 66.57 & 66.32 & 66.88 & 57.07 & 37 & 2 \\ \hline
\end{tabular}
\end{table}

\section{Marathi Subtask A: Identifying Hate, Offensive, and Profane Content from the Post}
\subsection{Data}

For the Marathi task, we used the original training and test sets provided by the organizers of the HASOC 2021. The whole dataset contains 2499 tweets, including: 1874 training and 625 test examples. The training set consists of 1205 texts of the NOT class and 669 texts of the HOF class. We used raw data as an input for our models. Following \cite{mishra2020multilingual,singh2020cfilt}, we experimented with the combination of the English, the Hindi, and the Marathi training sets provided by the organizers.

\subsection{Models}

We evaluated the following models:

\begin{itemize}
    \item XLM-RoBERTa$_{base}$ \cite{conneau2020unsupervised}, a transformer-based multilingual masked language model supporting 100 languages.
    \item LaBSE \cite{feng2020language}, a language-agnostic BERT sentence embedding model pre-trained on texts in 109 languages. 
\end{itemize}

\subsection{Experiments}

We experimented with the above-mentioned language models fine-tuned on monolingual and multilingual data. For model evaluation during the development phase, we used the random train and validation split in the ratio 80:20 with a fixed seed. We set the same model parameters as for English tasks.

\begin{table}[]
\caption{Model validation results for Marathi Subtask A, \%.}
\begin{tabular}{|l|l|l|l|}
\hline
Model & F1 & P & R \\ \hline
XLM-RoBERTa (Marathi dataset) &  83.87&  85.39&  83.39\\ \hline
XLM-RoBERTa (Marathi + Hindi) &  83.23&  83.82&  82.76\\ \hline
XLM-RoBERTa (Marathi + English) &  84.83&  85.03&  84.64\\ \hline
XLM-RoBERTa (Marathi + Hindi + English) &  84.35&  84.82&  83.95\\ \hline
LaBSE (Marathi) &  \textbf{87.76}&  87.82&  \textbf{87.68}\\ \hline
LaBSE (Marathi + Hindi) &  87.62&  \textbf{88.21}&  87.13\\ \hline
LaBSE (Marathi + English) &  87.62&  \textbf{88.21}&  87.13\\ \hline
LaBSE (Marathi + Hindi + English) &  86.34&  86.63&  86.08\\ \hline
\end{tabular}
\end{table}

Table 5 illustrates the results. It can be seen that LaBSE outperforms XLM-RoBERTa in all cases. Moreover, the F1-score of LaBSE fine-tuned only on the Marathi dataset are higher than the results of LaBSE fine-tuned on multilingual data. XLM-RoBERTa, on the other hand, mostly benefits from multilingual fine-tuning. 

For our final submission, we used a soft-voting ensemble of five LaBSE fine-tuned on the official Marathi dataset provided by the organizers of the competition. The results of this model on the test set are shown in Table 6.

\begin{table}[]
\caption{Performance of our final model for the Marathi Subtask A, official results, \%.}
\begin{tabular}{|l|l|l|l|l|l|l|}
\hline
\begin{tabular}[c]{@{}l@{}}F1 (our \\ model)\end{tabular} & \begin{tabular}[c]{@{}l@{}}F1 (winning\\ solution)\end{tabular} & \begin{tabular}[c]{@{}l@{}}P (our\\ model)\end{tabular} & \begin{tabular}[c]{@{}l@{}}P (winning\\ solution)\end{tabular} & Avg F1 & \begin{tabular}[c]{@{}l@{}}Number of\\ submitted\\ teams\end{tabular} & Rank \\ \hline
88.08 & 91.44 & 87.58 & 91.82 & 82.55 & 25 & 2 \\ \hline
\end{tabular}
\end{table}

\section*{Conclusion}

In this paper, we have presented the details about our participation in the HASOC Shared Task 2021. We have explored an application of domain-specific monolingual and multilingual BERT-based models to the tasks of binary and fine-grained classification of Twitter posts. We also proposed a one-vs-rest approach to discrimination between hate, offensive, and profane tweets. Further research can focus on analyzing the effectiveness of various text preprocessing techniques for harmful content detection and exploring how different transfer learning approaches can affect classification performance.

\section*{Acknowledgments}

The work on multi-label text classification was carried out by Anna Glazkova and supported by the grant of the President of the Russian Federation no. MK-637.2020.9.

\nolinenumbers

\bibliography{library}

\bibliographystyle{abbrv}

\end{document}